# Self-Supervised Pretraining Improves Cross-Site and Cross-Scale Robustness of Point Cloud Leaf-Wood Segmentation

**Heeju Mun[1], Tackang Yang[2,3*], Yunsoo Nam[1,3], Changhyun Choi[4]**

1 Department of Landscape Architecture and Rural Systems Engineering, Seoul National University, Seoul, South Korea

2 Interdisciplinary Program in Landscape Architecture, Seoul National University, South Korea

3 Integrated Major in Smart City Global Convergence, Seoul National University, Seoul, Republic of Korea

4 Research Institute of Agriculture and Life Sciences, Seoul National University, Seoul, Republic of Korea

*Corresponding author: Tackang Yang (didxorkd@snu.ac.kr)

## Abstract

The accuracy of existing leaf-wood segmentation methods for tree point clouds is inconsistent across forest types and site conditions. Self-supervised learning (SSL) on point clouds has been adopted to improve the generalization of deep learning models for forestry point cloud tasks, including biomass regression, individual tree segmentation and forest mapping. However, its applicability to leaf-wood segmentation remains untested. In this study, we pretrained Point-M2AE, a widely used SSL architecture for point clouds, on ShapeNet-55 augmented with 2,400 individual tree point clouds. For fine-tuning and inference, we employed voxel-based input partitioning during fine-tuning and recursive voxel subdivision at inference to handle the wide variation in point density across input point clouds, allowing the same model to operate at both individual-tree and plot scales without architecture change. Compared to the model without pretraining, the pretrained model improved wood IoU from 60.5% to 70.0% for needleleaf and from 69.7% to 76.3% for broadleaf trees. On a multi-site benchmark spanning four countries across three climatic zone (Germany, Finland, China, and Cameroon), the pretrained model achieved the smallest cross-site variation in wood IoU among compared methods (LeWos, CWLS, and PointTransformer) while attaining the highest overall mIoU (85.4%). Plot-level segmentation maintained accuracy comparable to individual-tree performance, with mIoU of 84.7% for broadleaf and 77.7% for needleleaf plots, demonstrating that the same model can be applied across scales without additional finetuning. As a downstream test in tropical forests, where dense canopies make leaf-wood segmentation particularly difficult, we applied our leaf-wood segmentation model and a subsequent quantitative structure model to estimate wood volume for 28 trees from Guyana, Indonesia, and Peru to assess whether the segmentation improvements achieved through SSL pretraining translate into improved downstream performance. The resulting volume estimates achieved the lowest error among all methods tested (MAE = 2.40 $m^3$), less than half that of algorithmic baselines (LeWos: 5.94 $m^3$; CWLS: 5.27 $m^3$). Our pretrained model thus provides a ready-to-use leaf-wood segmentation tool for TLS, and is publicly available at [https://github.com/heejumun/LeafWood_Segmentation.git]

# 1. Introduction

Terrestrial Laser Scanning (TLS) provides three-dimensional point clouds of forest, which is widely used to characterize tree structure and estimate aboveground biomass (AGB) (Åkerblom & Kaitaniemi, 2021; Dassot et al., 2011, 2012; Liang et al., 2016). At the individual-tree level, basic parameters such as diameter at breast height (DBH) and tree height can be extracted by simple geometric fitting (Calders et al., 2015; Liang et al., 2016). More complete characterization is obtained from quantitative structure models (QSM), which reconstruct full branching architecture by fitting hierarchical cylinder sets to the wood points (Du et al., 2019; Fan et al., 2020; Feng et al., 2024; Hackenberg et al., 2015; Raumonen et al., 2013). QSM-derived wood volume enables direct AGB estimation by multiplying wood density to the reconstructed volume, reducing the uncertainty associated with allometric equations based on DBH and height alone (Calders et al., 2015; Gonzalez de Tanago et al., 2018; Momo Takoudjou et al., 2018). However, the accuracy of structural assessment from point clouds ultimately depends on whether the woody structure is preserved as a clean and continuous representation within the point clouds.

Leaf-wood segmentation is a critical preprocessing step for structural assessment from point clouds, as imprecise separation can introduce noise or discontinuities into the predicted wood points and propagate errors into downstream QSM reconstruction. QSM pipelines can be categorized into skeleton-based methods and segmentation-based methods. Skeleton-based methods extract skeleton points that represent the central axes of tree structure (Du et al., 2019; Fan et al., 2020; Feng et al., 2024; Mei et al., 2017; Xu et al., 2007). Segmentation-based methods split woody point clouds into segments corresponding to the trunk and individual branches (Hackenberg et al., 2015; Liu et al., 2021; Raumonen et al., 2013). Both approaches then fit cylinder to each segment or skeleton part using local geometry along the branching hierarchy. The accuracy of cylinder fitting depends directly on the noise level and spatial continuity of the input woody points. Incorrect leaf-wood segmentation can distort local geometry and interrupt branch continuity, causing errors in fitted cylinder parameters such as radius and length (Feng et al., 2024; Mei et al., 2017). Chen et al. (2025) further demonstrated that leaf-wood segmentation errors translate into systematic over- or underestimation of QSM-derived whole tree volume and AGB. Accurate leaf-wood segmentation is therefore required for reliable woody volume and AGB estimation from point clouds.

Automated leaf-wood segmentation methods designed with handcrafted geometric features have shown limited cross-site transferability. The labor and time costs of manual leaf-wood segmentation have motivated extensive research on automated methods. Most existing automated approaches incorporate handcrafted geometric features into the segmentation pipeline. Algorithmic methods apply predefined thresholds to local geometric properties such as verticality, linearity, planarity, sphericity, and point connectivity (Ferrara et al., 2018; Tian & Li, 2022; Vicari et al., 2019; D. Wang et al., 2020; M. Wang & Wong, 2023). Machine learning-based methods learn nonlinear decision boundaries from annotated data using geometric features as input (Krishna Moorthy et al., 2020; D. Wang et al., 2017; Zhu et al., 2018), and more recent deep learning-based methods incorporate geometric features or network modules tailored to tree geometry with raw point coordinates and encode them jointly into high-dimensional representations (Bai et al., 2024; Dai et al., 2023; Jiang et al., 2023). Across all three categories, the reliance on handcrafted geometric features limits the broad applicability of leaf-wood segmentation models, because the ability of geometric features to distinguish leaf from wood points varies with point cloud characteristics such as density, occlusion, and noise that change across forest types and site conditions (Krishna Moorthy et al., 2020; Qin et al., 2019; Ren et al., 2022).

To reduce the dependence on handcrafted geometric features and improve transferability across forest sites, recent studies have adopted end-to-end learning on raw point coordinates for leaf-wood segmentation. End-to-end learning approaches learn representations directly from point coordinates and local neighborhood structure without requiring handcrafted geometric feature design (Hu et al., 2020; Qi et al., 2017; Thomas et al., 2019; Zhao et al., 2021). For leaf-wood segmentation task, Krisanski et al. (2021) showed that PointNet++ (Qi et al., 2017)-based segmentation generalizes across different forest types and sensor configurations without site-specific feature engineering, and Van den Broeck et al. (2025) applied PointTransformer (Zhao et al., 2021), KPConv (Thomas et al., 2019), and RandLA-Net (Hu et al., 2020) to TLS leaf-wood segmentation and reported consistent performance across tree sizes and species. However, because end-to-end learning is trained in a fully supervised manner, the performance is constrained by the size and diversity of available annotated data (Krisanski et al., 2021; Van den Broeck et al., 2025). Forest structural characteristics, from individual tree branching geometry to stand-level canopy complexity, vary widely across forest types, species compositions, understory and ground topography (Jo & Yang, 2024; Maeda et al., 2025), but leaf-wood annotated datasets inevitably cover only a limited subset of forest structural variability. Models trained on a limited annotated data tend to learn site-specific patterns rather than generalizable representations, leaving cross-site and cross-forest type generalization an open challenge (Sariyildiz et al., 2023; Shaheen et al., 2025; Zeng et al., 2023).

Self-supervised learning (SSL) enables learning point cloud representations from unannotated data by defining a pretext task, reducing the reliance on site-specific annotated data that limits supervised end-to-end learning of leaf-wood segmentation. For example, Masked Autoencoder (MAE) masks parts of the input and learns to reconstruct the masked regions, allowing the encoder to acquire latent representations without manual annotation (He et al., 2021; Pang et al., 2022). When the pretrained encoder is fine-tuned with annotated data, SSL-pretrained point cloud models, such as Point-BERT (Yu et al., 2022), Point-MAE (Pang et al., 2022), and MaskSurf (Y. Zhang et al., 2022), have been shown to outperform supervised learning models on downstream shape classification and part segmentation tasks (Ericsson et al., 2022; Zeng et al., 2023). SSL pretraining has since been extended to forest point cloud tasks including biomass regression and individual tree segmentation with consistent improvements over supervised training (Seely et al., 2026; Shaheen et al., 2025; Xiu et al., 2025). Seely et al. (2026) showed that an encoder pretrained on airborne LiDAR in a boreal forest reduced biomass regression error compared with training from scratch when fine-tuned on a temperate forest with different species composition, indicating that SSL-learned representations can generalize across forest sites. Other studies have reported that the degree of cross-domain transfer further depends on the similarity between the pretraining data and the target task domain (X. Chen & Cheng, 2025; Sariyildiz et al., 2023), raising the question of whether incorporating tree point clouds into pretraining would benefit leaf-wood segmentation. To our knowledge, however, SSL pretraining has not yet been examined for TLS leaf-wood segmentation, and the effect of adding tree point clouds to the pretraining data on cross-site segmentation transferability remains untested.

In this study, we investigate whether end-to-end learning with SSL pretraining on tree point clouds can improve leaf-wood segmentation accuracy and reduce domain dependency of the model across sites. Specifically, we make three contributions: (1) we apply SSL pretraining for TLS leaf-wood segmentation for the first time by pretraining Point-M2AE on synthetic objects augmented with tree point clouds as target domain data, (2) we evaluate the cross-site generalization of the SSL-pretrained model against algorithmic and learning-based baselines across six benchmark sites, and (3) we evaluate the suitability of the predicted wood points for downstream QSM-based volumetric estimation by comparing QSM-derived volumes against destructively measured reference volumes at three tropical sites.

# 2. Method

## 2.1. Data

| Scale | Dataset | Size | Source | Purpose |
|---|---|---|---|---|
| - | ShapeNet-55 | 57,448 objects | Chang et al. (2015) | • SSL pretraining |
| Individual Tree-level | KR-BL | 2,199 trees | AI-Hub (2023) | • SSL pretraining<br>• Leaf-wood segmentation fine-tuning |
| | KR-NL | 201 trees | | |
| | SWDEU-BL | 6 trees | Weiser et al. (2024) | • Individual tree-level leaf-wood segmentation evaluation |
| | SWDEU-NL | 4 trees | | |
| | DEU-BL | 6 trees | Harry et al. (2024) | |
| | FIN-NL | 12 trees | | |
| | NCHN-BL | 15 trees | Wan et al. (2021) | |
| | NCHN-NL | 15 trees | | |
| | GUY | 10 trees | Gonzales de Tanago et al. (2018) | • TreeQSM-based tree volume estimation evaluation |
| | IND | 10 trees | | |
| | PER | 9 trees | | |
| Plot-level | CAM-BL-PL | 1 plot (15 trees) | Harry et al. (2024) | • Plot-level leaf-wood segmentation evaluation |
| | FIN-NL-PL | 1 plot (15 trees) | | |

**Table 1. Dataset used for training and evaluation of leaf-wood segmentation model.** Dataset names follow the convention: country code – leaf type (BL = broadleaf, NL =needleleaf), with the suffix -PL appended for plot-level data.

To bridge the domain gap between synthetic ShapeNet objects and natural tree geometry, we augmented ShapeNet-55 (Chang et al., 2015) with KR datasets for SSL pretraining. ShapeNet-55 contains 57,448 objects including 55 object categories. Each object is represented by 8,192 points uniformly sampled from synthetic mesh surface. The KR-BL and KR-NL datasets consist of individual 2,199 broadleaf and 201 needleleaf trees from 27 broadleaf and 9 needleleaf species, each manually segmented from TLS scans collected across five urban sites in South Korea (Daegu, Dongtan, Jeju, Sejong, and Wonju) using Leica BLK360 scanner (AI-Hub, 2023). The species diversity and clean individual tree segmentation provide varied and high-quality tree geometry to the pretraining pool alongside synthetic ShapeNet objects. Each tree point cloud, ranging from approximately $10^4$ to over $10^7$ points, was down-sampled to 8,192 points using farthest point sampling (FPS) to match the ShapeNet input size while retaining the global tree shape. All pretraining samples consisted solely of point coordinates without class or segmentation labels, as Point-M2AE learns representations by reconstructing masked subsets of the input point coordinates.

To partition large point clouds in the KR dataset into fixed-size input units for fine-tuning and inference, we applied voxel-based spatial partitioning. Global down-sampling to 8,192 points, as used for pretraining, would discard the local geometric detail that point-wise leaf-wood classification depends on. Spatial partitioning instead divides the point clouds into volumetric units, preserving local geometry for fine-tuning and inference (Bai et al., 2024). During fine-tuning, the batch size requirement constrains input to 8,192 points per forward pass. Therefore, each tree point cloud was partitioned into eight spatial voxels (2 × 2 × 2), and 8,192 points were randomly sampled from each voxel for supervised training. During inference, batch size was set to one, and all points within each voxel were processed without random sampling. Under GPU memory constraints, each voxel had to remain within the memory limit, so the input point cloud was recursively partitioned by dividing each voxel into eight smaller voxels along the three coordinate axes until every voxel contained fewer points than a predefined maximum threshold. However, segmentation quality varies with the combination of maximum threshold and patch configuration (patch size and number of patches). The optimal combination was therefore determined empirically based on segmentation accuracy, as described in Appendix (Figure A2).

To construct a benchmark dataset for evaluating leaf-wood segmentation across different sites and spatial scales, we collected leaf-wood annotated TLS point clouds from three published datasets, spanning six sites and two spatial scales (Table 1). The six sites cover temperate, boreal, and tropical forests across four countries, with two of the six sites also used for plot-level evaluation. SWDEU-BL and SWDEU-NL were obtained from Weiser et al. (2024), scanned in temperate broadleaf and needleleaf stands in southwest Germany with a RIEGL VZ-400i, containing 7 species. DEU-BL, FIN-NL, CAM-BL-PL, and FIN-NL-PL were obtained from Harry et al. (2024), scanned in a temperate broadleaf forest in Germany, a boreal needleleaf forest in Finland, and a tropical broadleaf forest in Cameroon, all with a RIEGL VZ-400i. NCHN-BL and NCHN-NL were obtained from Wan et al. (2021), scanned in temperate broadleaf and needleleaf forest in northern China with a RIEGL VZ-1000. Because annotation quality directly affects the interpretability of point-wise evaluation, we excluded trees that exhibited any of the following issues by visual inspection: (1) severe sparsity, where the woody structure was not recoverable even visually and thus could not serve as a reliable reference, (2) inconsistent leaf-wood labels, where clearly woody structures were labeled as leaf, and (3) incomplete tree points, where parts of the target tree were missing or points from neighboring trees were included. Representative excluded cases are shown in Appendix (Figure A1). After screening, we retained 58 individual trees and two forest plots across six sites, each from a different site (Table 1). This collection provides variation in point density and structural representation, and therefore enable evaluation of leaf-wood segmentation under heterogeneous acquisition conditions (Hackenberg et al., 2015; Yan et al., 2025).

To evaluate QSM-based woody volume estimation from segmented wood points against destructively measured reference volumes, we used the publicly available tropical tree TLS dataset of Gonzalez de Tanago et al. (2018). Dataset provides TLS point clouds and destructively measured reference volumes simultaneously for 29 trees across three tropical sites, Guyana (GUY, 10 trees), Indonesia (IND, 10 trees), and Peru (PER, 9 trees) (Gonzalez de Tanago et al., 2018). Each tree was scanned with a RIEGL VZ-400 prior to destructive harvest. To obtain a single per-tree reference volume in m$^3$ from these site-specific measurements, we aggregated the recorded components as follows. For GUY, the woody volume was estimated as the sum of four components: stump, stem, buttresses, and branches. The stump was modeled as an elliptical cylinder using the minor and major basal diameters. The stem above 1 m was calculated as a stack of frustums between 1 m interval diameter measurements, using the truncated-cone formula:

$$V_{frustum} = \frac{\pi L}{12}(D_1^2 + D_1 D_2 + D_2^2) \quad (1)$$

Where L is the segment length and $D_1$, $D_2$ are the diameters at the two endpoints. Buttresses were modelled as triangular plates with $V = \frac{1}{2} L \cdot W \cdot H$ from the recorded length, width, and height.

Branches were calculated as frustums from the recorded base and top diameters and segment length. For and IND, the original data provides volumes of each components (stem, branch, twig, leaf, stump, fruit), and we summed all aboveground wood fractions per tree. For all three sites, the total per-tree reference volume serves as the validation target for the QSM-based volume estimation in Section 2.3.

## 2.2. Leaf-Wood Segmentation

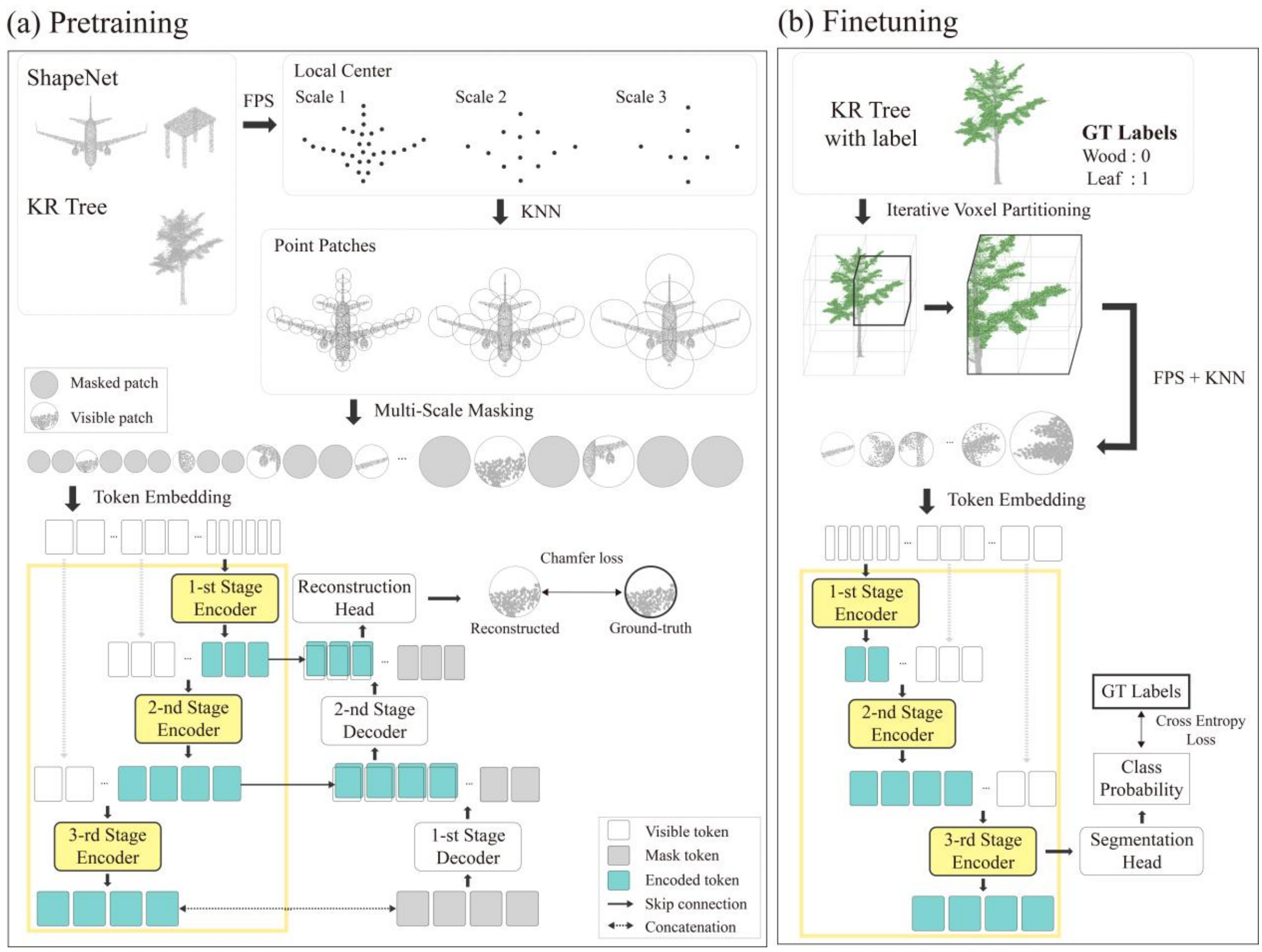

**Figure 1. Overview of the proposed pipeline** during (a) pretraining and (b) finetuning stage.

To enable the encoder to learn representations from tree point clouds without manual annotation, we selected Point-M2AE as our backbone (He et al., 2021; Pang et al., 2022). MAE-based models learn representations from unannotated data by masking a large portion of the input and training the encoder to reconstruct the masked parts. This pretext task allows the encoder to learn point cloud representations from unannotated TLS data, which is expected to improve generalization across sites not represented in the annotated training data (Krishna Moorthy et al., 2020; Van den Broeck et al., 2025). The encoder processes local point patches constructed by farthest point sampling (FPS) and k-nearest neighbors (KNN) rather than absolute coordinates or fixed grids, which allows it to handle the variation in point density across different TLS acquisition settings without domain-specific modifications such as predefined geometric thresholds or tree-specific modules (Bai et al., 2024; Dai et al., 2023). Among MAE-based point cloud SSL methods, Point-M2AE (R. Zhang et al., 2022) incorporates multi-scale masking with a hierarchical encoder-decoder architecture and reported the strongest downstream performance on tasks including object classification and part segmentation tasks.

To investigate whether pretraining on tree point clouds improves the encoder's transferability to leaf-wood segmentation, we pretrained Point-M2AE (R. Zhang et al., 2022) under two data configurations: (1) ShapeNet-55 (Chang et al., 2015) alone, and (2) ShapeNet-55 augmented with KR-NL and KR-BL data. We retained the original Point-M2AE architecture and hyperparameters, which were reported to perform well across multiple downstream tasks in the original study (R. Zhang et al., 2022). Both configurations were pretrained for 300 epochs using the Adam optimizer (learning rate 0.001, weight decay 0.05) with a cosine learning rate scheduler (CosLR). For the loss function, we used Chamfer Distance loss $L_{CD}$ (Equation 1) between the reconstructed point set $S_{gt}$ and ground-truth point set $S_{rec}$ in the masked regions:

$$L_{CD}(S_{gt}, S_{rec}) = \frac{1}{|S_{gt}|}\sum_{x \in S_{gt}} \min_{y \in S_{rec}} \|x - y\|_2^2 + \frac{1}{|S_{rec}|}\sum_{y \in S_{rec}} \min_{x \in S_{gt}} \|x - y\|_2^2 \quad (1)$$

Here, $|\cdot|$ denotes the number of points in a set and $\|\cdot\|_2$ is the Euclidean norm. For each configuration, we split the data into training and validation sets with a 7:3 ratio. We selected the best checkpoint based on object classification accuracy on the validation set, evaluated by a support vector machine (SVM) classifier trained on features extracted from the frozen encoder, following the Point-M2AE model selection protocol (R. Zhang et al., 2022).

To adapt the pretrained encoder to leaf-wood segmentation, we attached a segmentation head to Point-M2AE encoder and fine-tuned the full model on annotated KR-BL and KR-NL data. The segmentation head follows the Point-M2AE part segmentation design: it propagates the three-scale encoder features to per-point representations via PointNet++ feature propagation (Qi et al., 2017) and maps them to per-point leaf and wood probabilities through linear projection layers and a softmax. We fine-tuned the model under three encoder initialization settings: (1) Randomly-initialized (without pretraining), (2) ShapeNet-pretrained, and (3) ShapeNet + KR-pretrained, to isolate the effect of SSL pretraining and the additional benefit of including tree data in pretraining dataset. We split the KR dataset into training, validation, and test sets with a 7:2:1 ratio and fine-tuned model with all three initialization settings for 300 epochs using the same setting as pretraining. For each training iteration, voxel blocks served as the input unit, from which 8,192 points were randomly sampled. We used the cross-entropy loss $L_{CE}$ (Equation 2) computed over all labeled points:

$$L_{CE} = -\frac{1}{N}\sum_{i=1}^{N}\sum_{c=1}^{C} y_{i,c} \log(\hat{y}_{i,c}) \quad (2)$$

where N is the total number of points in a batch, C is the number of classes (C = 2, leaf and wood), $y_{i,c}$ is the ground-truth label for point i and class c, and $\hat{y}_{i,c}$ is the predicted class probability. We selected the checkpoint with the highest validation accuracy.

To assign a label to every point at inference step, we processed each iteratively partitioned voxel (Section 2.1) through the network without subsampling. Because each inference voxel contains more points than the training input, we enlarged the patch configuration accordingly. The optimal threshold and patch configuration were jointly determined as described in Appendix (Figure A2).

## 2.3. Model Evaluation

To assess whether self-supervised pretraining improves leaf-wood segmentation, we compared the segmentation performance of models fine-tuned under three initialization settings (Randomly-initialized, ShapeNet-pretrained, and ShapeNet + KR-pretrained) on the KR test set (19 needleleaf and 221 broadleaf trees). We quantitatively evaluated performance using four metrics defined in Equations (3)-(6): overall accuracy (OA), mean Intersection over Union (mIoU), leaf IoU, and wood IoU. We also visually inspected segmentation outputs to identify which regions of the tree point cloud drove the metric differences among models fine-tuned under the three initialization settings.

$$\mathrm{OA} = \frac{1}{N}\sum_{i=1}^{N}(y_i = \hat{y}_i) \tag{3}$$

$$\mathrm{mIoU} = \frac{1}{C}\sum_{c=1}^{C}\frac{TP_c}{TP_c + FP_c + FN_c} \tag{4}$$

$$\mathrm{Leaf\ IoU} = \frac{TP_{leaf}}{TP_{leaf} + FP_{leaf} + FN_{leaf}} \tag{5}$$

$$\mathrm{Wood\ IoU} = \frac{TP_{wood}}{TP_{wood} + FP_{wood} + FN_{wood}} \tag{6}$$

To evaluate the model under plot-level conditions without relying on a preceding individual tree segmentation step, we additionally evaluated the three initialization settings on two forest plots (CAM-BL-PL and FIN-NL-PL). Unlike individual tree-level leaf-wood segmentation, plot-level leaf-wood segmentation operates directly on raw point clouds containing overlapping crowns and neighboring trees, providing a more realistic assessment of model performance under operational conditions. Following the same approach as at the individual tree level, performance was quantified using the four metrics defined in Equations (3)-(6) and segmentation outputs were visually inspected.

To evaluate the cross-site generalization of SSL-pretrained models for leaf-wood segmentation, we compared the performance of three Point-M2AE-based models (Randomly-initialized, ShapeNet-pretrained, and ShapeNet + KR-pretrained) against representative baselines on the multi-site benchmark dataset described in Section 2.1. We selected baselines to represent the two main methodological categories in TLS leaf-wood segmentation: algorithmic and learning-based. For the algorithmic approaches, we chose LeWos (D. Wang et al., 2020) and CWLS (Yan et al., 2025), which separate leaf and wood using geometric features and point connectivity. LeWos is a widely adopted algorithmic method with publicly available code, and CWLS is a recent method that has reported strong performance across multiple tree species. For the learning-based category, we selected a PointTransformer (Van den Broeck et al., 2025; Zhao et al., 2021) fine-tuned for leaf-wood segmentation. This method represents the end-to-end learning approach most directly comparable to our models, differing primarily in its pretraining strategy. All baselines were run with their default or recommended settings without additional hyperparameter tuning: threshold value 0.125 for LeWos, the default parameter set for CWLS, and the official inference configuration for the PointTransformer. We evaluated all models on the 58 trees (31 needleleaf and 27 broadleaf) across three sites using the quantitative point-wise metrics (Equation (3)-(6)). The cross-site consistency of model was quantified as the standard deviation of site-level mean values for each metric, computed separately for needleleaf and broadleaf trees, where lower standard deviation indicates more consistent performance across sites.

To assess whether the model predicted wood points are suitable for downstream QSM-based volumetric reconstruction, we estimated the whole-tree woody volume from the outputs of all six methods (LeWos, CWLS, PT, and three Point-M2AE-based models) and compared the estimated volumes against the destructive reference volume defined in Section 2.1. For each tree and each of six methods, we extracted the predicted wood points and fitted a hierarchical cylinder using TreeQSM v2.0 using its default cover set parameters (PatchDiam1 = 1, PatchDiam2Min = 1, PatchDiam2Max =1) (Raumonen et al., 2013). Because TreeQSM produces slightly different cylinder fits between runs due to its random initialization, we ran it ten times for every tree and method combination and used the mean of the ten runs as the per-tree volume estimate. One tree from the PER site (MDD01_006) was excluded from the comparison because its fitted QSM of manually extracted wood showed an exceptionally large standard deviation across ten runs (SD = 2.70 $m^3$), about 3.6 times the mean SD of 0.75 $m^3$ across all trees, indicating that the cylinder reconstruction was unstable independently of any segmentation method. We quantified agreement between the QSM-derived whole-tree woody volume $V_i'$ and the destructive reference $V_i$ using the mean absolute error (MAE; Equation (7)) computed across the 28 trees, and we additionally reported per-site MAE for GUY, IND, and PER. Reconstructed QSMs were visually inspected to examine how differences in predicted wood points were reflected in QSM reconstruction and whether these differences were associated with changes in volumetric estimation.

$$\mathrm{MAE} = \frac{\sum_{i=1}^{n} |V_i \; - \; V_i'|}{n} \tag{7}$$

# 3. Results

## 3.1. Effect of SSL Pretraining on Leaf-Wood Segmentation

Self-supervised pretraining improved leaf-wood segmentation performance compared to Randomly-initialized model on the KR test set, with the largest gains appearing in wood prediction (Table 2). All three encoder configurations achieved overall accuracy above 90% for both needleleaf and broadleaf trees. However, the differences among the three configurations concentrated in wood IoU. For needleleaf trees, wood IoU increased monotonically from 60.5% (Randomly-initialized) to 67.2% (ShapeNet-pretrained) and 70.0% (ShapeNet + KR-pretrained), while leaf IoU remained stable (92.9-93.6%). For broadleaf trees, wood IoU increased from 69.7% to 72.2% and then to 76.3%. ShapeNet + KR-pretrained model further improved OA (90.5% to 92.2%), mIoU (79.8% to 82.9%), and leaf IoU (87.4% to 89.5%) beyond the ShapeNet-pretrained model.

**Table 2. Leaf-wood segmentation results by leaf type for three different pretraining configuration.** The best results for each leaf type are highlighted in bold.

| Leaf type | Model | OA | mIoU | Leaf IoU | Wood IoU |
|---|---|---|---|---|---|
| **Needleleaf** | Randomly-initialized | 93.6 | 76.7 | 92.9 | 60.5 |
| | ShapeNet-pretrained | 93.8 | 80.1 | 92.9 | 67.2 |
| | ShapeNet + KR-pretrained | **94.5** | **81.8** | **93.6** | **70.0** |
| **Broadleaf** | Randomly-initialized | 90.4 | 78.8 | 87.8 | 69.7 |
| | ShapeNet-pretrained | 90.5 | 79.8 | 87.4 | 72.2 |
| | ShapeNet + KR-pretrained | **92.2** | **82.9** | **89.5** | **76.3** |

Visual inspection showed that the performance gains from pretraining were driven by improved wood points extraction in regions where leaf and wood points are densely intermingled (Figure 2). All three configurations successfully segmented main stems and major branches, but differed in how much woody structure was preserved where leaf and wood points overlap in the upper canopy. For broadleaf trees, the Randomly-initialized model frequently misclassified woody as leaf on the branches and branch junctions (Figure 2a, b). ShapeNet-pretrained model reduced these wood-to-leaf errors on branches but wood-to-leaf misclassification at branch junctions persisted (Figure 2c, d). ShapeNet + KR-pretrained model further reduced wood-to-leaf errors on both branches and branch junctions (Figure 2e, f). For needleleaf trees, the Randomly-initialized and ShapeNet-pretrained model missed major branches embedded within dense leaf points (Figure 2g, i), whereas ShapeNet + KR-pretrained model extracted many of the missed branches (Figure 2k) and recovered additional fine-scale woody structures in regions with densely clustered needle shoots (Figure 2h, j, l). For both broadleaf and needleleaf trees, ShapeNet + KR-pretrained model reduced wood-to-leaf errors with a slight increase in leaf-to-wood errors (Figure 2e, f, l). Despite the trade-off, the reduction in wood-to-leaf errors was larger than the increase in leaf-to-wood errors, and therefore ShapeNet + KR-pretrained model achieved the highest wood IoU across both leaf types (Table 2).

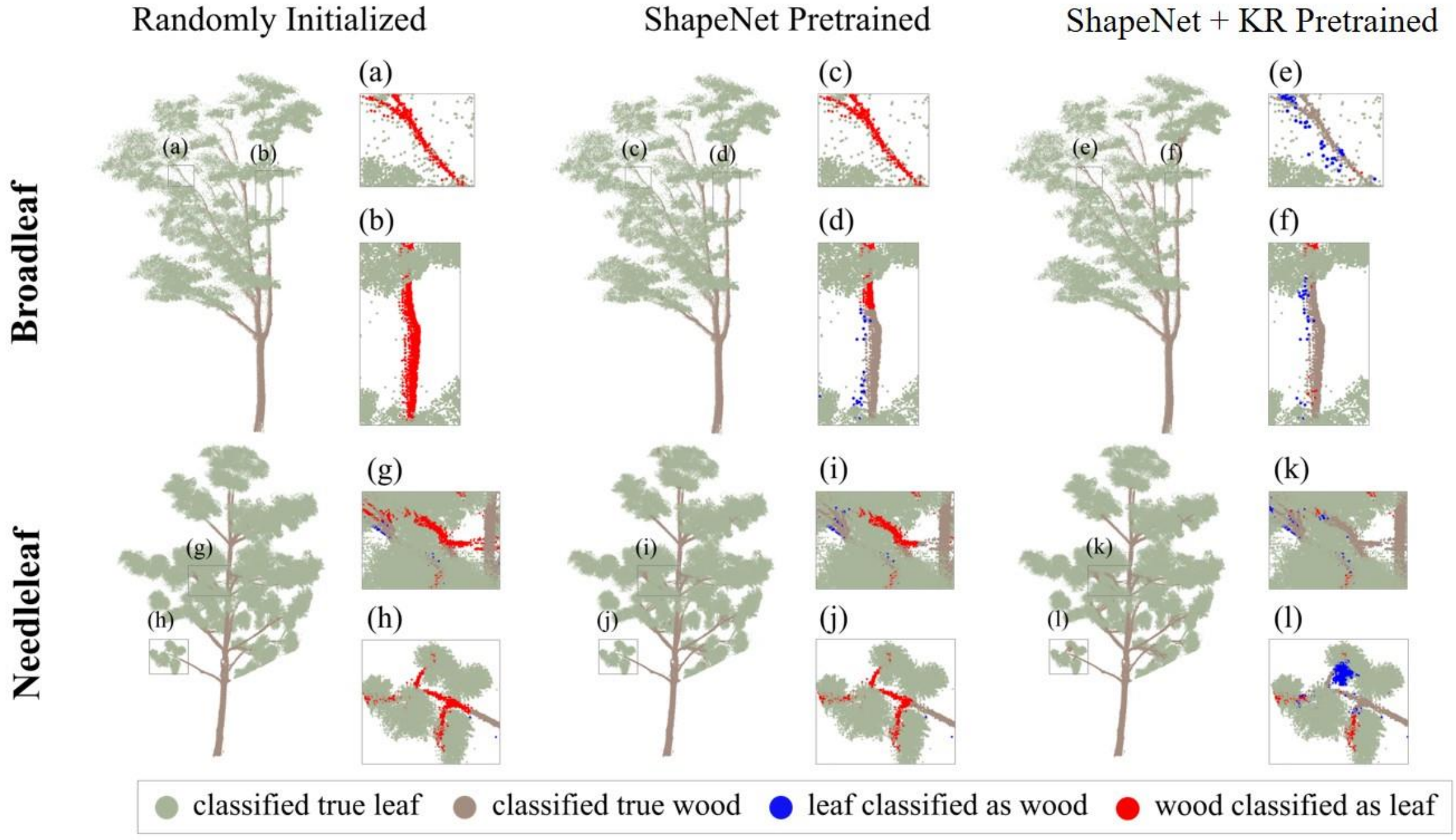

**Figure 2. Visual Comparison of leaf-wood segmentation results on the KR test set**. Top row: broadleaf tree with Randomly-initialized (a,b), ShapeNet-pretrained (c,d), and ShapeNet + KR-pretrained (e,f) encoders. Bottom row: needleleaf tree with Randomly-initialized (g,h), ShapeNet-pretrained (i,j), and ShapeNet + KR -pretrained (k,l) encoders. Zoomed panels show regions where leaf and wood points overlap.

### 3.2. Cross-Site Leaf-Wood Segmentation Performance

Across all 58 trees at six sites, ShapeNet + KR-pretrained model achieved the highest performance across all four metrics (OA, mIoU, leaf IoU, and wood IoU) among all compared methods (Table 3). ShapeNet + KR-pretrained model attained OA of 94.3%, mIoU of 85.4%, leaf IoU of 92.3%, and wood IoU of 78.5% (Table 3). Among the three Point-M2AE-based models, ShapeNet-pretrained model and ShapeNet + KR-pretrained model outperformed all baselines across all four metrics. Segmentation accuracy improved monotonically with pretraining data configuration, consistent with the pattern observed at the KR test set.

**Table 3. Overall leaf-wood segmentation performance across the full benchmark (six sites),** averaged across all trees including needleleaf and broadleaf. The best results for each leaf type are highlighted in bold.

| | OA | mIoU | Leaf IoU | Wood IoU |
|---|---|---|---|---|
| **LeWos** | 91.2 | 78.5 | 88.0 | 69.0 |
| **CWLS** | 92.8 | 82.3 | 89.5 | 75.1 |
| **PT** | 93.6 | 83.0 | 91.6 | 74.4 |
| **Randomly-initialized** | 93.3 | 82.3 | 91.2 | 73.4 |
| **ShapeNet-pretrained** | 93.9 | 84.3 | 91.9 | 76.7 |
| **ShapeNet + KR-pretrained** | **94.3** | **85.4** | **92.3** | **78.5** |

Cross-site consistency of leaf-wood segmentation performance was most apparent in wood IoU (Figure 3). For broadleaf trees, the median wood IoU of ShapeNet + KR-pretrained model ranged from 80.4% to 81.8% across sites, with a spread of only 1.4%, whereas LeWos ranged from 49.6% to 85.7% with a spread of 36.1% (Figure 3b). For needleleaf trees, the median wood IoU of ShapeNet + KR-pretrained model ranged from 74.3% to 81.8% compared with 54.3%-75.4% (LeWos), 68.8%-80.8% (CWLS), 65.2%-82.9% (PT), 66.6%-76.0% (Randomly-initialized), and 71.2% to 79.6% (ShapeNet-pretrained) (Figure 3a). ShapeNet + KR-pretrained model exhibited the most consistent segmentation performance across sites for both broadleaf and needleleaf with the smallest standard deviation in all four evaluation metrics (Figure 4). Among the three Point-M2AE-based models, the standard deviation of all four metrics monotonically decreased from Randomly-initialized model to ShapeNet-pretrained model to ShapeNet + KR-pretrained model.

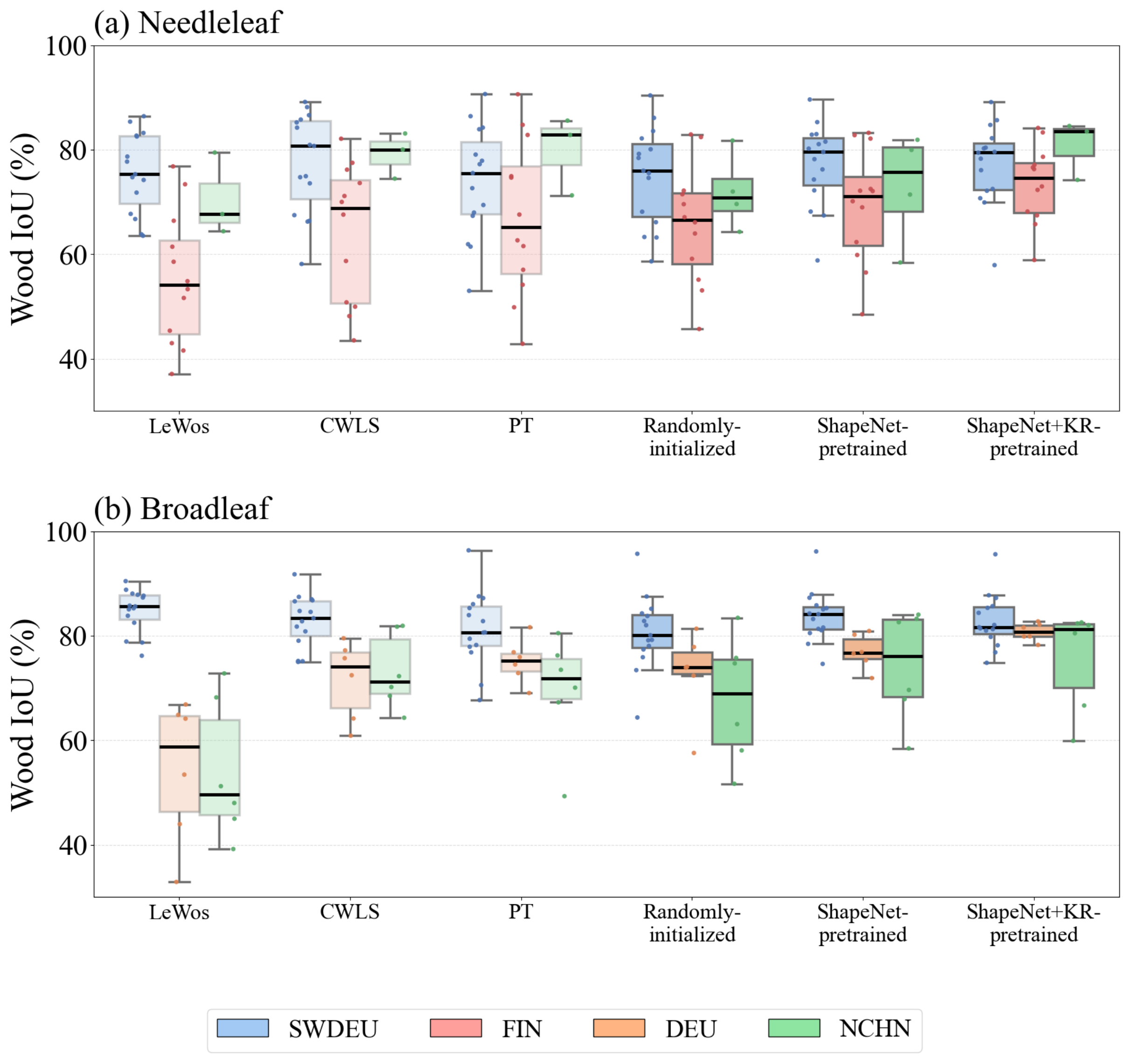

**Figure 3. Per-tree wood IoU across six sites and six methods** for (a) needleleaf and (b) broadleaf trees. Methods include LeWos, CWLS, PointTransformer, and three Point-M2AE-based models (Randomly-initialized, ShapeNet-pretrained, ShapeNet + KR-pretrained).

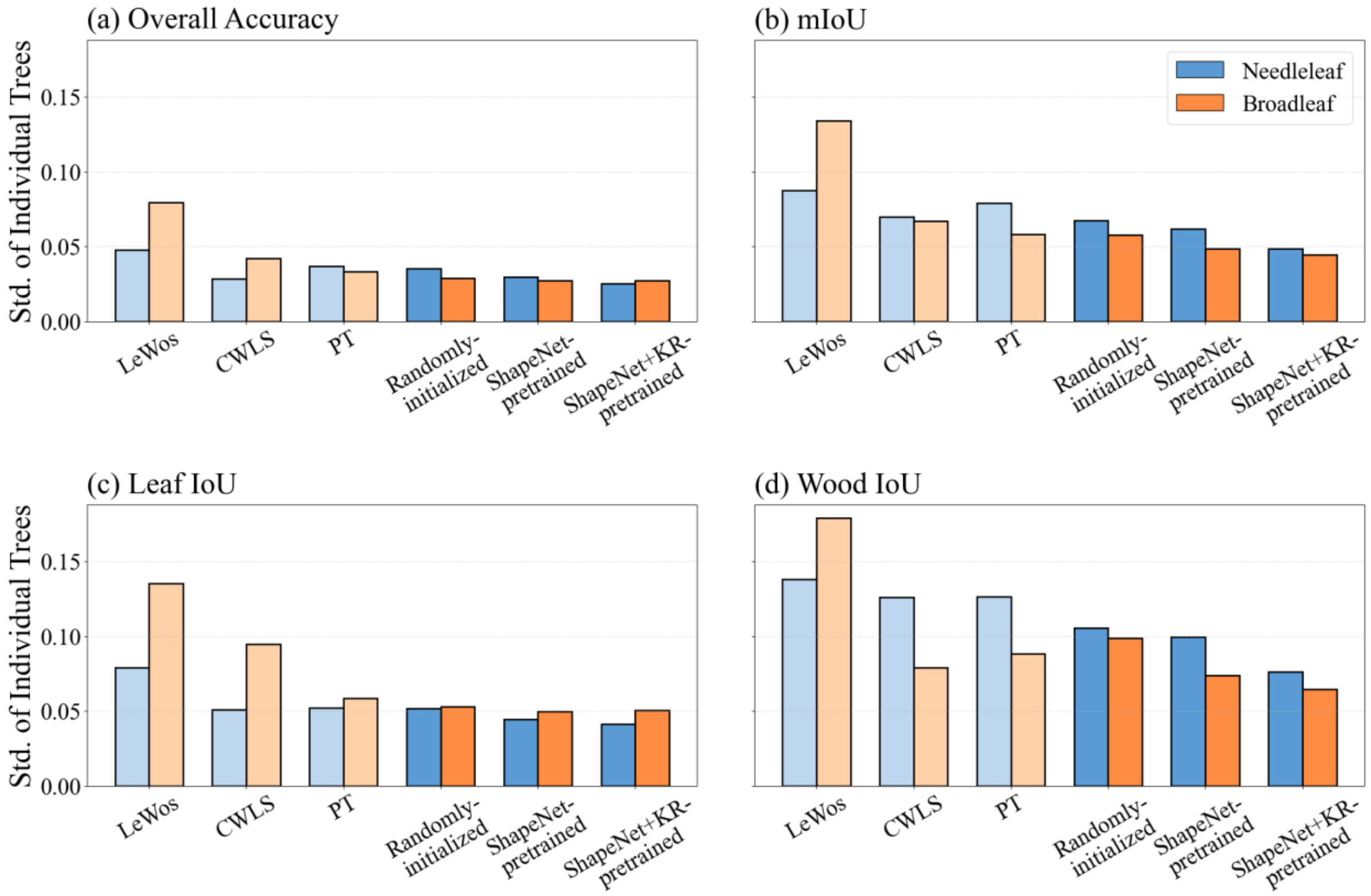

**Figure 4. Standard deviation of individual tree-level metrics** across six methods for needleleaf and broadleaf trees. Lower values indicate more consistent performance across sites. (a) Overall Accuracy, (b) mIoU, (c) Leaf IoU, (d) Wood IoU.

### 3.3. Scalability from Individual Trees to Plot-Level Point Clouds

At the plot-level, SSL pretraining improved leaf-wood segmentation performance, similar to the improvements observed at the individual tree-level (Table 4, Figure 5). For the needleleaf plot, ShapeNet-pretrained model achieved the highest values across all four metrics, whereas the ShapeNet + KR-pretrained model showed the highest performance for the broadleaf plot. Both pretrained models outperformed the Randomly-initialized model for both leaf types. However, unlike at the individual tree-level, incorporating the KR dataset into pretraining reduced segmentation performance for the needleleaf plot (FIN-NL-PL) relative to the ShapeNet-pretrained model (Table 4). Visual inspection supported why the broadleaf and needleleaf plots responded differently to the inclusion of the KR dataset during SSL pretraining. For the broadleaf plot (CAM-BL-PL), the ShapeNet + KR-pretrained model recovered fine branches in the upper canopy, consistent with the improvement in segmentation performance from ShapeNet-pretrained model to ShapeNet + KR-pretrained model. For the needleleaf plot (FIN-NL-PL), the increase in leaf-to-wood misclassification from ShapeNet-pretrained model to ShapeNet + KR-pretrained model, which was already evident at the individual tree-level, became more pronounced at the plot-level, resulting in lower segmentation performance for the ShapeNet + KR-pretrained model.

Compared with the individual tree-level, plot-level segmentation exhibited a different overall performance pattern. Wood IoU decreased across all three configurations for both leaf types. In contrast, the response of leaf IoU depended on leaf type: leaf IoU decreased in needleleaf plot (FIN-NL-PL) but increased in the broadleaf plot (CAM-BL-PL) relative to the corresponding individual tree-level results (Table 2 and 4). The reduction in leaf IoU for the needleleaf plot can be attributed to the plot structure, where dense shoot clusters from adjacent trees overlapped. This overlapping placed leaf and wood points from multiple trees within the same input voxels (Figure 5), reducing leaf-wood segmentation accuracy in the upper canopy.

**Table 4. Quantitative evaluation of plot level leaf-wood segmentation** for needleleaf and broadleaf plots. The best results for each leaf type are highlighted in bold.

| Leaf type | Model | OA | mIoU | Leaf IoU | Wood IoU |
|---|---|---|---|---|---|
| **Needleleaf** | Randomly-initialized | 90.6 | 70.2 | 89.6 | 50.9 |
| | ShapeNet-pretrained | **92.9** | **78.5** | **91.8** | **65.2** |
| | ShapeNet + KR-pretrained | 92.4 | 77.7 | 91.2 | 64.3 |
| **Broadleaf** | Randomly-initialized | 93.2 | 77.3 | 92.4 | 62.1 |
| | ShapeNet-pretrained | 94.4 | 81.3 | 93.5 | 69.1 |
| | ShapeNet + KR-pretrained | **95.3** | **84.7** | **94.6** | **74.8** |

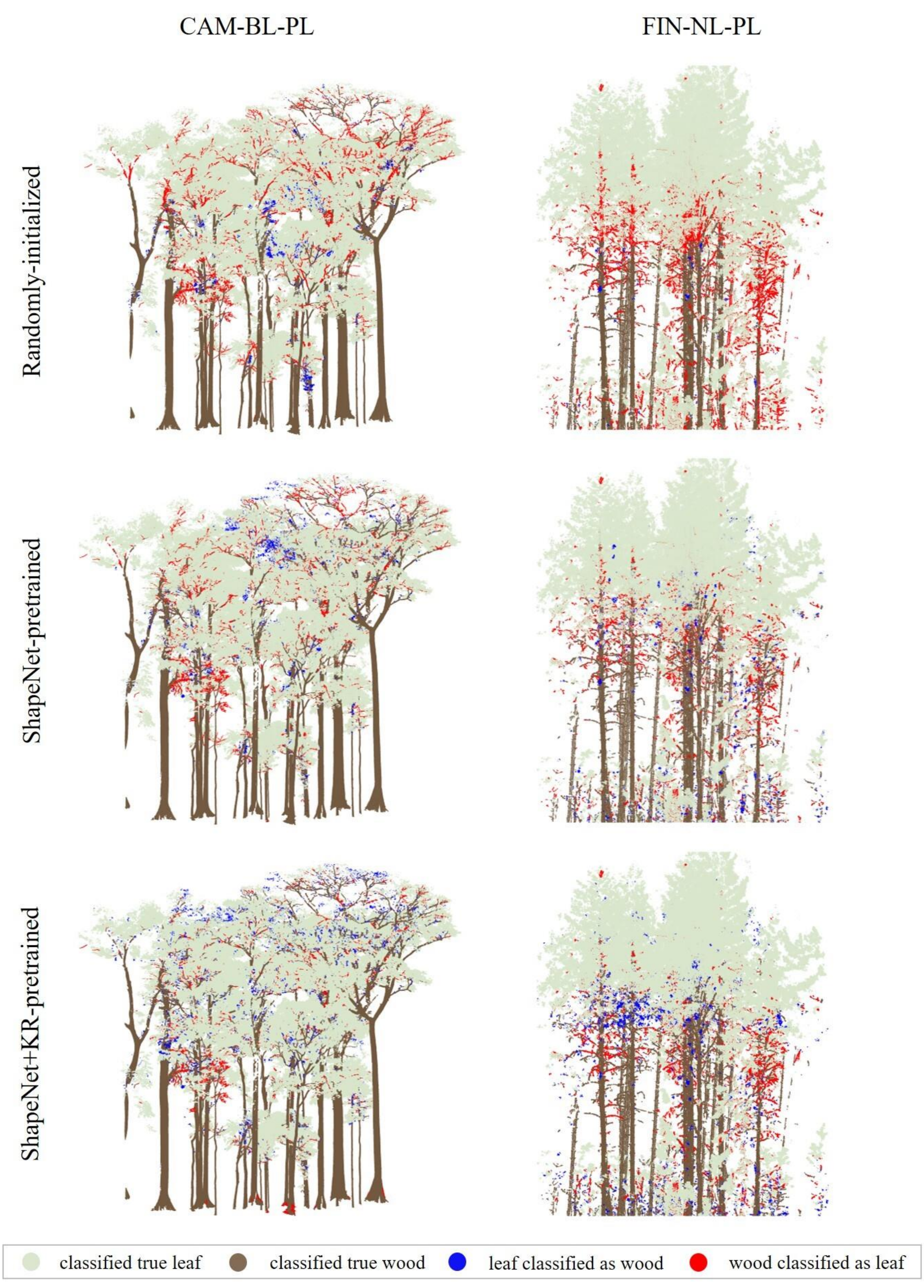

**Figure 5. Visual results of plot level leaf-wood segmentation** for needleleaf (top) and broadleaf (bottom) plots. Leaf: ground truth leaf-wood labels. Right: predicted leaf-wood labels

### 3.4. Volumetric Estimation from QSM Fitting

Across all 28 tropical trees from three sites (GUY: 10 trees, IND: 10 trees, PER: 8 trees), ShapeNet + KR-pretrained model achieved the lowest overall MAE (2.40 $m^3$), followed by ShapeNet-pretrained model (2.45 $m^3$), PT (2.73 $m^3$), Randomly-initialized model (3.67 $m^3$), CWLS (5.27$m^3$), and LeWos (5.94 $m^3$). Among the three Point-M2AE-based models, SSL pretraining reduced volume estimation error relative to the Randomly-initialized model. The overall MAE decreased from 3.67 $m^3$ for Randomly-initialized model to 2.45 $m^3$ for ShapeNet-pretrained model and 2.40 $m^3$ for ShapeNet + KR-pretrained model. At the site level, ShapeNet + KR-pretrained model achieved the lowest MAE at GUY (3.92 $m^3$), ShapeNet-pretrained model achieved the lowest MAE at PER (2.56 $m^3$), and PT achieved the lowest MAE at IND (0.61 $m^3$). For GUY sites, estimated volumes were consistently below the 1:1 line across all six methods. LeWos and CWLS showed larger deviations from the 1:1 line at GUY than the three Point-M2AE-based models and PT. At IND and PER, estimated volumes were distributed on both sides of the 1:1 line, with no consistent under- or overestimation pattern across methods. Relative MAE remained below 32% for ShapeNet-pretrained model, ShapeNet + KR-pretrained model, and PT at all three sites, whereas LeWos and CWLS exceeded 40% at both GUY and PER.

**Table 5. Evaluation of volumetric estimation** MAE ($m^3$) and relative MAE (%) of QSM-derived total tree volume against the destructive reference. The lowest MAE per column is highlighted.

| Method | All sites | GUY (n=10) | IND (n=10) | PER (n=8) |
|---|---|---|---|---|
| | MAE ($m^3$) | MAE ($m^3$) | MAE ($m^3$) | MAE ($m^3$) |
| LeWos | 5.94 | 7.00 (50.2%) | 1.55 (21.1%) | 9.26 (43.5%) |
| CWLS | 5.27 | 7.14 (50.0%) | 1.30 (17.7%) | 7.36 (44.2%) |
| PT | 2.73 | 4.40 (30.1%) | **0.61 (10.0%)** | 3.19 (17.0%) |
| Randomly-initialized | 3.67 | 5.63 (39.3%) | 0.79 (10.9%) | 4.59 (25.3%) |
| ShapeNet-pretrained | 2.45 | 4.09 (27.2%) | 0.70 (10.3%) | **2.56 (13.4%)** |
| ShapeNet + KR-pretrained | **2.40** | **3.92 (25.7%)** | 0.72 (11.3%) | 2.61 (15.6%) |

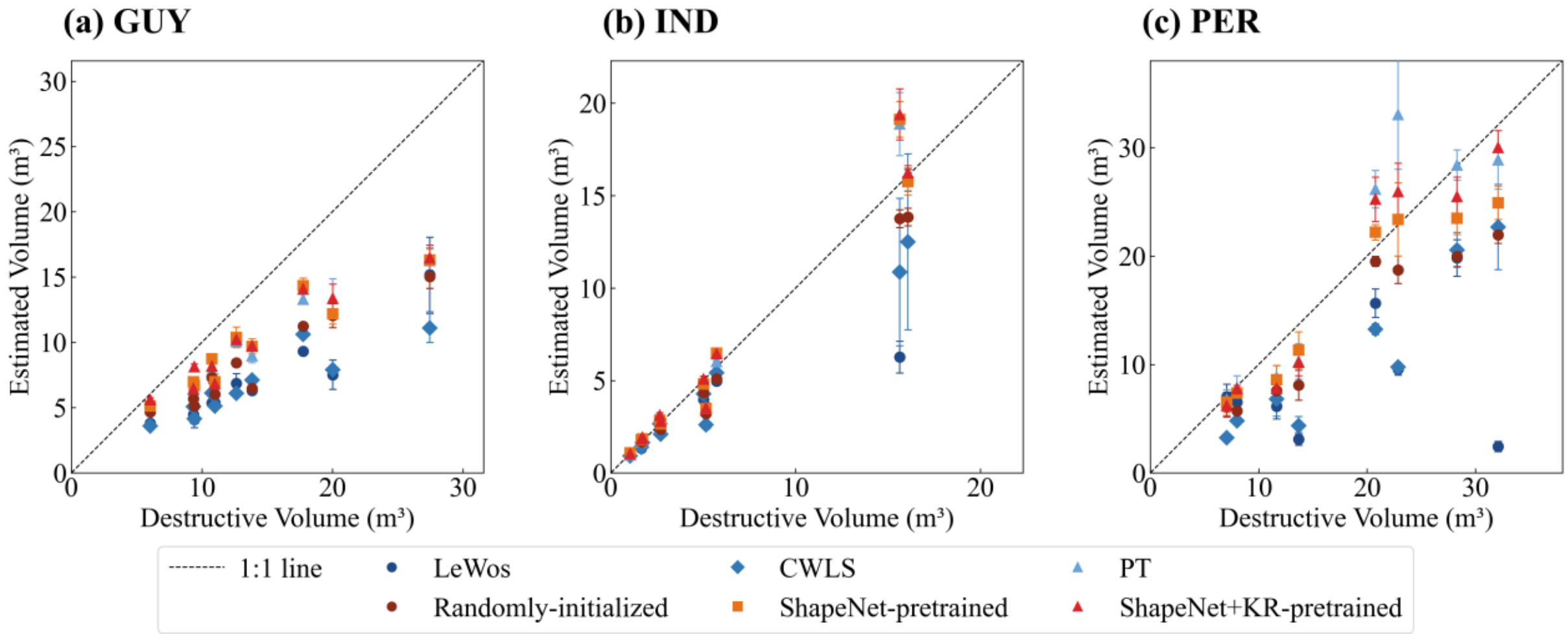

**Figure 6. Comparison of QSM-derived total tree volume against destructive reference measurement** for six leaf-wood segmentation methods (LeWos, CWLS, PT, Radnomly-initialized, ShapeNet-pretrained, ShapeNet + KR-pretrained) at three sites (GUY: Guyana, IND: Indonesia, PER: Peru). Points represent the mean of 10 QSM iterations per tree, with error bars indicating ±1 SD, due to randomized initial parameters in QSM.

Visual inspection of trees for which ShapeNet-pretrained model produced lower volumetric estimation error than ShapeNet+KR-pretrained model revealed differences in the reconstruction of fine branch structures (Figure 7). ShapeNet+KR-pretrained model recovered additional fine branches that were absent from the ShapeNet-pretrained model outputs. For IND07, ShapeNet+KR-pretrained model produced an estimated volume of 16.82 $m^3$ compared with 16.39 $m^3$ from ShapeNet-pretrained model against a reference volume of 16.08 $m^3$, resulting in a higher estimation error. A similar pattern was observed for PER08, where ShapeNet+KR-pretrained model estimated 25.09 $m^3$ compared with 22.91 $m^3$ from ShapeNet-pretrained model against a reference volume of 20.76 $m^3$. In both examples, the additional fine branches recovered by ShapeNet+KR-pretrained model were associated with larger volume estimates and higher volumetric estimation error than ShapeNet-pretrained model.

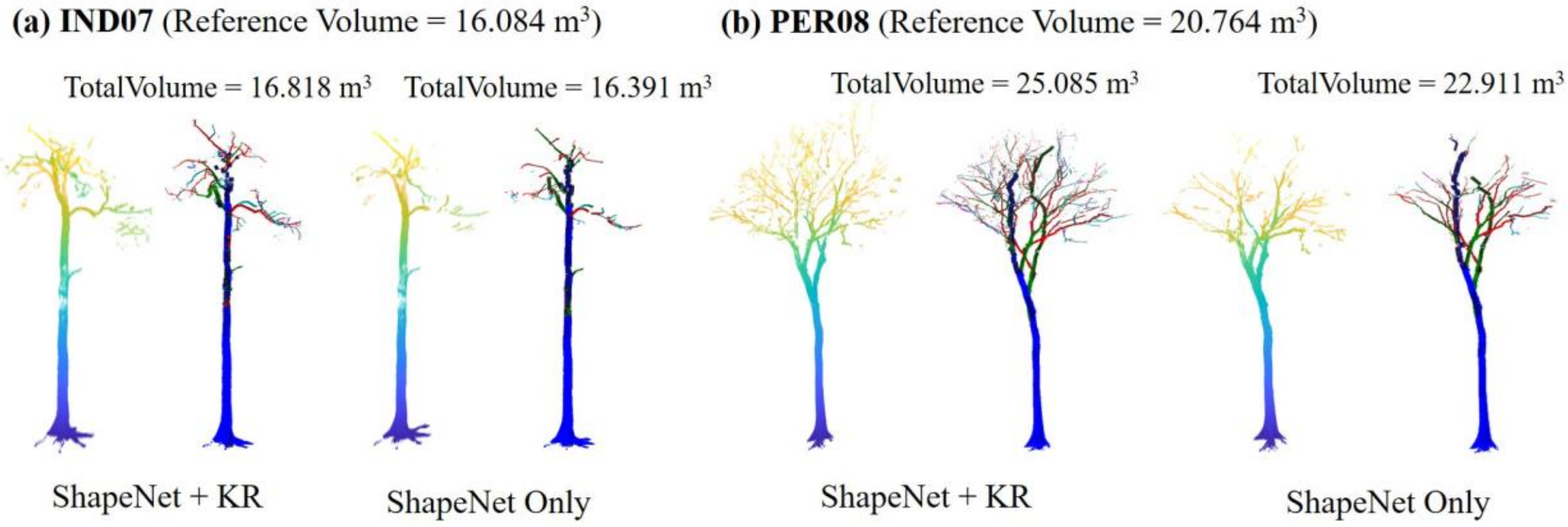

**Figure 7. Example of QSM reconstructions from ShapeNet + KR-pretrained and ShapeNet-pretrained wood points for two representative trees.** (a) IND07 (reference volume = 16.08 $m^3$) and (b) PER08 (reference volume = 20.76 $m^3$).

# 4. Discussion

## 4.1. The Pretrained Encoder Transfers to Leaf-Wood Segmentation Across Forest Sites

The encoder pretrained through the masked reconstruction pretext task transferred successfully to leaf-wood segmentation. Wood IoU increased according to the pretraining data configuration on both the KR test set and the cross-site benchmark (Tables 2 and 3), and visual inspection confirmed more accurate wood point extraction in fine branches and inner crown regions as pretraining progressed from Randomly-initialized model to ShapeNet-pretrained model to ShapeNet + KR-pretrained model (Figure 2). The magnitude of wood IoU gain depended more on the relevance of the pretraining data to the leaf-wood segmentation task than on the volume of data added to pretraining. Adding the KR dataset increased the pretraining data by only 4% (2,400 trees added to 57,448 objects), yet the wood IoU gain from ShapeNet-pretrained model to ShapeNet + KR-pretrained model exceeded half the wood IoU gain from Randomly-initialized model to ShapeNet-pretrained model. Deep learning models are generally expected to improve logarithmically with pretraining data volume (Sun et al., 2017), so 4% increase in data would predict a correspondingly small gain in performance. The wood IoU gain from ShapeNet-pretrained model to ShapeNet + KR-pretrained model, observed here instead matches recent findings that downstream task performance depends more on the relevance of pretraining data to the target task domain than on data volume alone (X. Chen & Cheng, 2025; Sariyildiz et al., 2023). The wood IoU gain was not matched by a proportional gain in leaf IoU, which remained stable across the three configurations. In some cases, the increase in wood IoU coincided with increased leaf-to-wood misclassification (Figure 2), suggesting that pretraining shifted the decision boundary toward preserving woody structures rather than uniformly reducing classification error in both classes.

The encoder trained on the KR dataset was also applied to the unseen cross-site benchmark without additional fine-tuning. The performance improvement observed on the KR test set also appeared on the cross-site benchmark, where the ShapeNet + KR-pretrained model achieved the highest overall accuracy and the lowest cross-site variation among all compared methods (Table 3, Figure 4). Cross-site variation decreased from Randomly-initialized model to ShapeNet-pretrained model to ShapeNet + KR-pretrained model across all four evaluation metrics (Figure 4). Seely et al. (2026) (Seely et al., 2026) demonstrated a related form of transfer in biomass regression task, where an encoder pretrained on boreal forest was fine-tuned on a temperate forest to confirm generalization across forest types. The present study extends this generalization evidence by applying the pretrained and fine-tuned encoder directly to the unseen cross-site benchmark without additional fine-tuning, showing consistent performance gains across sites not represented during pretraining or fine-tuning.

### 4.2. SSL Pretraining Improves QSM-Based Volumetric Estimation

Wood points predicted by the SSL pretrained models produced lower overall volumetric error than Randomly-initialized model. The overall MAE across 28 tropical trees at three sites decreased monotonically from Randomly-initialized model to ShapeNet-pretrained model to ShapeNet + KR-pretrained model, with the ShapeNet + KR-pretrained model achieving the lowest MAE (2.40 $m^3$) among all evaluated methods including the algorithmic and learning-based baselines (Table 5, Figure 6). The reduction in volumetric error is consistent with the wood IoU gain reported in Section 4.1, and with previous studies showing that leaf-wood segmentation errors propagate into QSM reconstruction and biomass estimation (S. Chen et al., 2025; Van den Broeck et al., 2025).

The ShapeNet + KR-pretrained model improved the recovery of fine branch structures and inner crown region in QSM reconstruction, but this improvement did not consistently reduce volumetric error. Visual comparison showed that the ShapeNet + KR-pretrained model recovered additional fine branches beyond branches captured by the ShapeNet-pretrained model (Figure 7). However, captured fine branches produced higher error for some trees at IND and PER, because QSM cylinders fitted to fine branches are prone to overestimation against a destructive reference that excludes branches below 10 cm in diameter (Table 5, Figure 7) (Gonzalez de Tanago et al., 2018; Morales & MacFarlane, 2025). Volumetric error also arose independently of segmentation quality at the GUY site. All six methods underestimated volumes relative to the destructive reference (Figure 6) and the same underestimation persisted when manually segmented ground-truth wood points were used for QSM reconstruction (Appendix Figure A3). Because underestimation occurred even when wood points contained no leaf-wood segmentation error, the point clouds at GUY likely lacked complete woody structure independent of segmentation quality. Wang et al. (2020) (D. Wang et al., 2020) attributed a similar pattern in large tropical trees to upper canopy occlusion limiting cylinder fitting, indicating that point cloud completeness constrains volumetric accuracy independently of how well leaf and wood points are classified.

### 4.3. Implications and Limitations

The improved wood point extraction of fine branches and branch junctions in inner crown regions (Figure 2) addresses a challenge that has consistently limited the accuracy of leaf-wood segmentation methods. Inner crown regions have been identified in previous studies as the primary source of leaf-wood classification difficulty because wood and leaf points share similar geometrical properties and overlapping branches obscure class boundaries (S. Chen et al., 2025; D. Wang et al., 2020). SSL pretraining on the ShapeNet + KR dataset improved wood point extraction by preserving these fine branches, visible both on the KR test set and in the QSM reconstructions of tropical trees used for volumetric evaluation (Figure 2 and 7). Fine branches are difficult to represent accurately in TLS-derived QSMs and difficult to measure destructively (Morales & MacFarlane, 2025; Morhart et al., 2024). The additional recovery of fine branches enabled by the inclusion of the KR dataset in pretraining indicates that wood points predicted by the ShapeNet + KR-pretrained model can support the reconstruction of fine woody structures that are difficult to quantify through destructive volume measurements.

The highest accuracy and the lowest cross-site variation achieved by the ShapeNet + KR-pretrained model across the six benchmark sites (Table 3, Figure 4) support the use of the pretrained model on diverse and unseen forest conditions without additional site-specific fine-tuning. The benchmark used in the present study covered a broader range of forest conditions than most previous leaf-wood segmentation studies, which have focused on tropical forests (Van den Broeck et al., 2025; D. Wang et al., 2020) or compared multiple forest types within a single country (Dai et al., 2023; Yan et al., 2025). The model weights, code, and pretraining configurations are publicly available for further evaluation on additional forest sites.

The lowest overall MAE among all evaluated methods achieved by the ShapeNet + KR-pretrained model at the three tropical sites (Table 5) supports the use of SSL pretrained leaf-wood segmentation model for volume estimation in forest conditions where destructive reference data are limited and difficult to obtain. QSM-derived volumes have served as ground truth for allometric equations when destructive measurement is not feasible (Brede et al., 2022; Yang et al., 2025). Large tropical trees are representative example where destructive measurement is difficult, as they are often underrepresented in destructive harvest datasets, and uncertainty in allometric biomass estimation is often large (Calders et al., 2015; Gonzalez de Tanago et al., 2018). The volumetric evaluation in this study was conducted using large tropical trees with destructive reference measurements, and both SSL-pretrained models reduced volumetric error relative to the Randomly-initialized model (Table 5). This result suggests the applicability of predicted wood points from SSL-pretrained leaf-wood segmentation to TLS-based volume estimation of large tropical trees.

Several limitations remain. The KR tree point clouds were always combined with ShapeNet during pretraining, so the independent contribution of tree point clouds on leaf-wood segmentation performance improvement cannot be isolated from the combined effect of the two datasets. Forest types such as subtropical and dry tropical forests remain unrepresented in the cross-site benchmark, because publicly available TLS point clouds with leaf-wood annotations are unavailable for these conditions, and expanding such benchmarks would enable more comprehensive evaluation of cross-site generalization. In addition, the volumetric evaluation was conducted only on broadleaf tropical trees, and has not been evaluated in coniferous forests, where leaf-wood segmentation remains necessary because leaf-off conditions are unavailable, so whether the same reduction in volumetric error extends to coniferous forests remains untested.

## Data and code availability

The code and SSL pretrained encoder weights (ShapeNet-pretrained and ShapeNet + KR-pretrained), and leaf-wood segmentation fine-tuned model weights for each configuration are publicly available at https://github.com/heejumun/LeafWood_Segmentation.git. The KR tree point cloud dataset is openly accessible at AI Hub (https://aihub.or.kr/aihubdata/data/view.do?dataSetSn=71458).

LeWos algorithm is available at https://github.com/dwang520/LeWoS. PT algorithm is available at https://github.com/qforestlab/leaf-wood-segmentation-with-deep-learning. CWLS algorithm is available at https://zenodo.org/records/17303206.

The NCHN benchmark site data is available at https://doi.org/10.5061/dryad.rfj6q5799. The FIN, DEU, and FIN-NL-PL, CAM-BL-PL site data is available at https://doi.org/10.5281/zenodo.13285640. The SWDEU benchmark site data is available at https://doi.org/10.11588/data/UUMEDI.

## CRediT authorship contribution statement

**Heeju Mun:** Writing – original draft, Writing – review & editing, Visualization, Validation, Conceptualization, Formal analysis, Investigation, Methodology. **Tackang Yang:** Writing – review & editing, Conceptualization, Data curation, Formal analysis, Methodology, Supervision. **Yunsoo Nam:** Writing – review & editing, Conceptualization, Supervision. **Changhyun Choi:** Writing – review & editing, Conceptualization, Supervision.

## Declaration of competing interest

The authors declare that they have no known competing financial interests or personal relationships that could have appeared to influence the work reported in this paper.

## Acknowledgements

This research was supported by Korea Environment Industry & Technology Institute (KEITI) through Technology Development Project for the Creation and Management of Ecosystem-based Carbon Sinks, funded by Korea Ministry of Climate, Energy and Environment (MCEE) (202300218237). We thank Youngryel Ryu for his support on this manuscript.

## Appendix

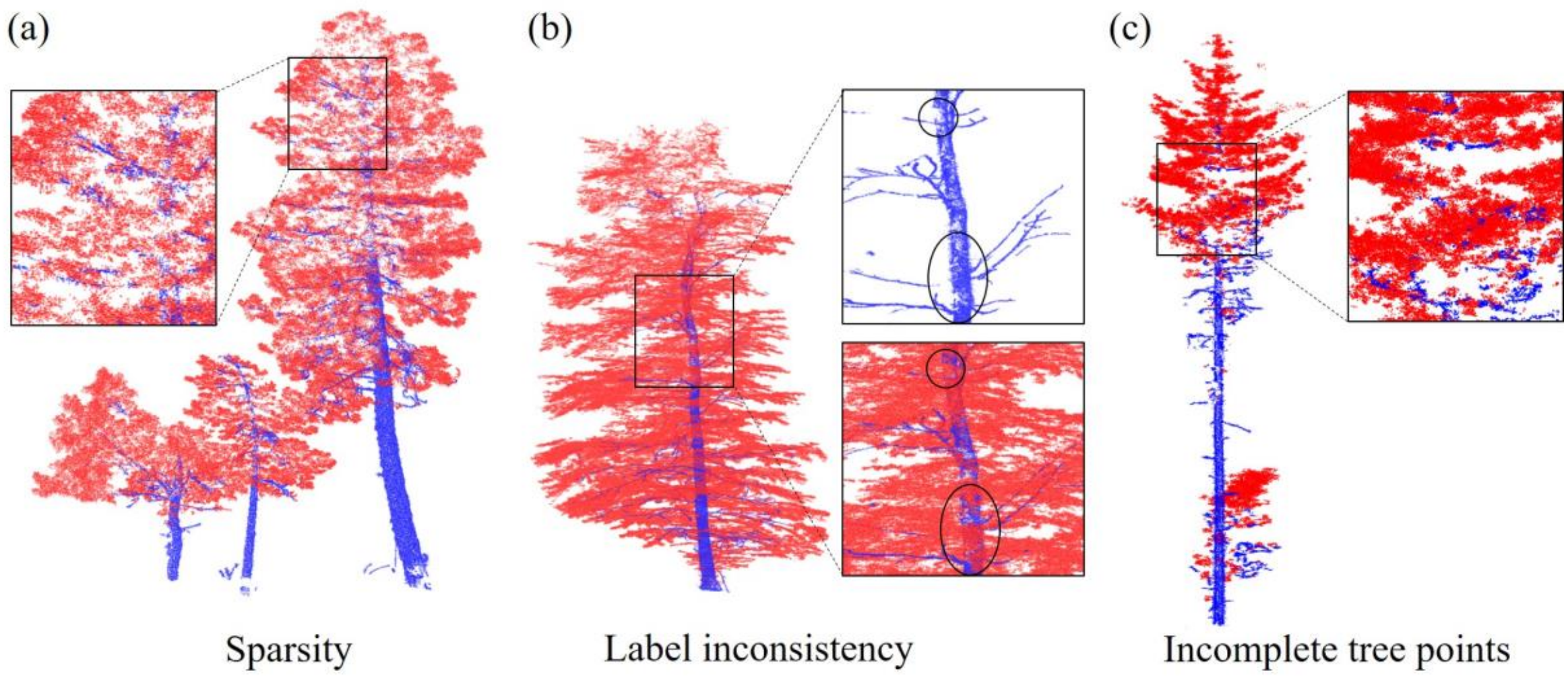

**Figure A1. Representative excluded cases from the candidate trees of the three source datasets.** Wood points are shown in blue and leaf points in red. (a) Sparsity: the woody structure is too sparsely sampled to serve as a reliable reference. (b) Label inconsistency: too many woody components within the crown are labeled as leaf. (c) Incomplete tree points: large portions of the upper crown and finer branches are missing. Such trees were excluded from the multi-site benchmark prior to evaluation.

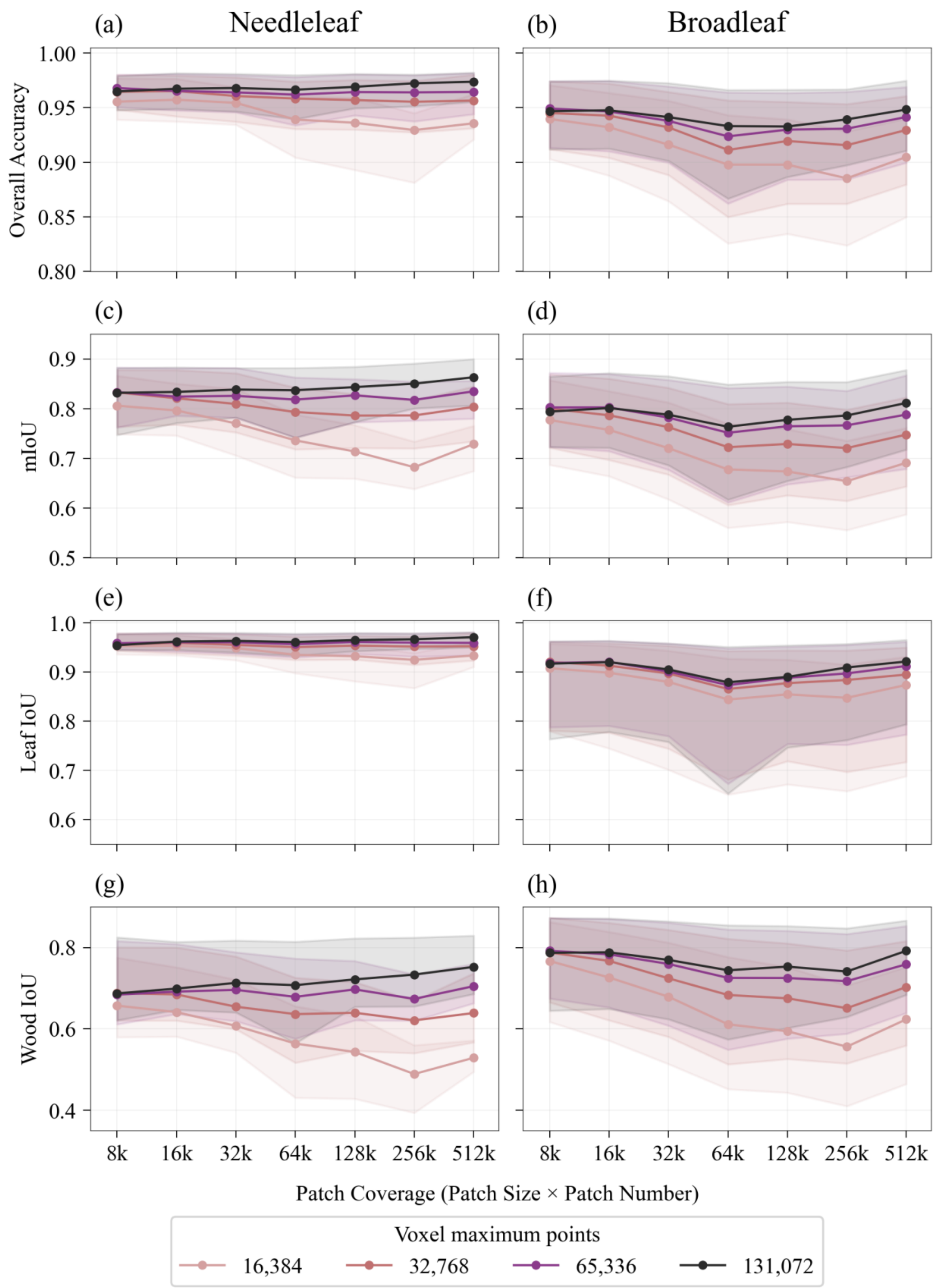

**Figure A2. Joint optimization of voxel maximum point threshold and patch coverage on**

**segmentation performance.** Each metrics are validated on test set: (a), (b) overall accuracy, (c), (d) mean IoU, (e), (f) leaf IoU, and (g), (h) wood IoU for needleleaf (left column) and broadleaf (right column) trees. The x-axis shows patch coverage, defined as the product of patch size and patch number. Line colors correspond to four voxel maximum point thresholds: 16,384, 32,768, 65,336, and 131,072. Shaded bands indicate the standard deviation across trees.

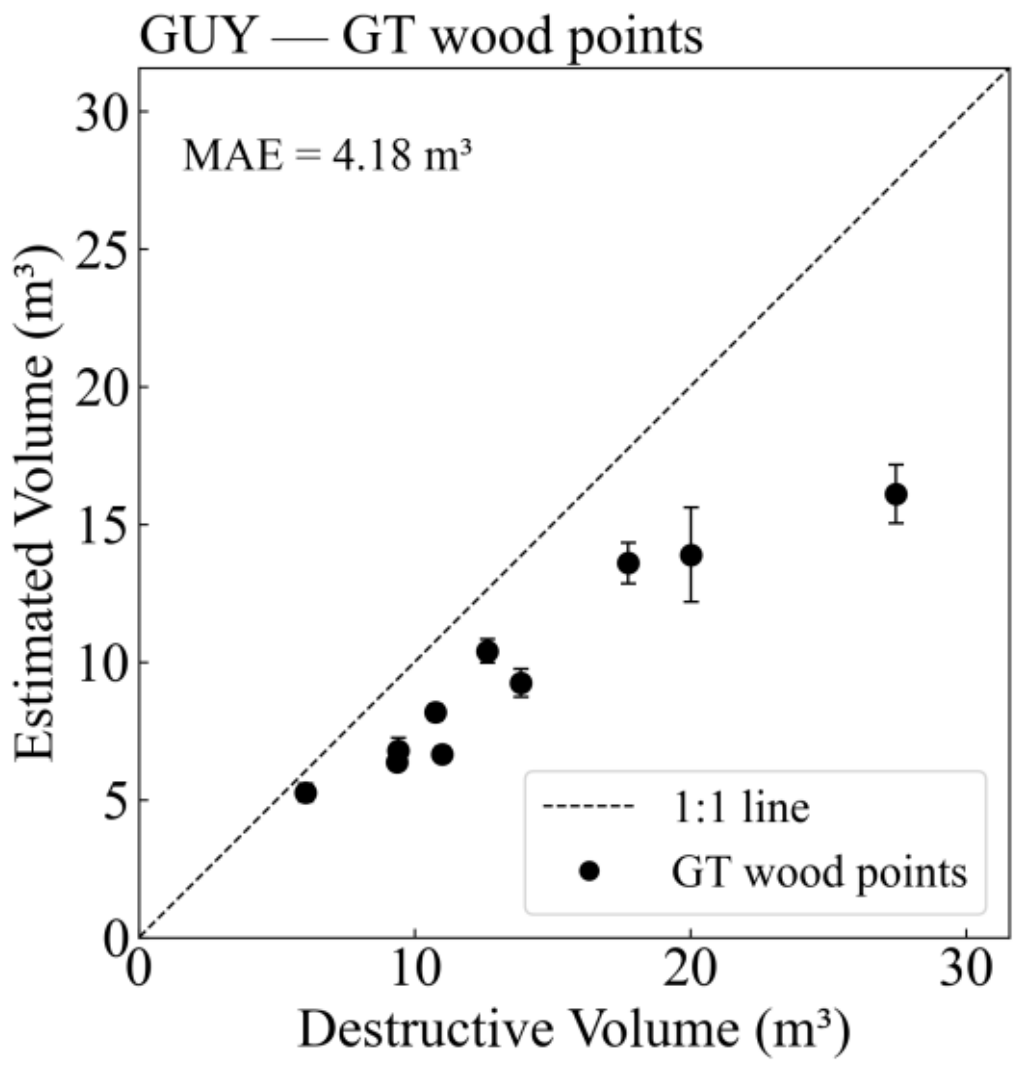

**Figure A3. QSM-derived volume from manually segmented (GT) wood points against destructive reference measurement** for 10 trees at GUY**.** Points and error bars as in Figure 6.